\pdfoutput=1

\documentclass[11pt]{article}

\usepackage[final]{acl}

\usepackage{times}
\usepackage{latexsym}

\usepackage{amssymb}
\usepackage{amsmath}
\usepackage{float} 
\usepackage{algorithm}
\usepackage{listings}
\usepackage{parskip}
\usepackage{booktabs} 
\usepackage{multirow}
\usepackage{hyperref}
\usepackage{float}

\usepackage[T1]{fontenc}

\usepackage[utf8]{inputenc}

\usepackage{microtype}

\usepackage{inconsolata}

\usepackage{graphicx}

\title{DropLoRA: Sparse Low-Rank Adaptation for Parameter-Efficient Fine-Tuning}

\author{Haojie Zhang \\
  \texttt{tayeechang@gmail.com} \\}

\begin{document}
\maketitle
\begin{abstract}
LoRA-based large model parameter-efficient fine-tuning (PEFT) methods use low-rank decomposition to approximate updates to model parameters. However, compared to full-parameter fine-tuning, low-rank updates often lead to a performance gap in downstream tasks. To address this, we introduce DropLoRA, a novel pruning-based approach that focuses on pruning the rank dimension. Unlike conventional methods that attempt to overcome the low-rank bottleneck, DropLoRA innovatively integrates a pruning module between the two low-rank matrices in LoRA to simulate dynamic subspace learning. This dynamic low-rank subspace learning allows DropLoRA to overcome the limitations of traditional LoRA, which operates within a static subspace. By continuously adapting the learning subspace, DropLoRA significantly boosts performance without incurring additional training or inference costs. Our experimental results demonstrate that DropLoRA consistently outperforms LoRA in fine-tuning the LLaMA series across a wide range of large language model generation tasks, including commonsense reasoning, mathematical reasoning, code generation, and instruction-following. Our code is available at \url{https://github.com/TayeeChang/DropLoRA}.
\end{abstract}

\section{Introduction}

Large language models (LLMs) have demonstrated remarkable proficiency in diverse cognitive tasks spanning machine translation, information extraction, question answering, human-like dialogue systems, and logical reasoning \cite{guo2025deepseek, achiam2023gpt, brown2020language}. While this methodology typically involves two-phase training - pre-training on extensive datasets followed by instruction-based fine-tuning (IFT) for downstream task optimization. The substantial computation and memory overhead required for effective instruction fine-tuning pose significant barriers to implementing these architectures in resource-constrained scenarios \cite{grattafiori2024llama}. Consequently, Efficient fine-tuning techniques based on models are increasingly gaining popularity and attention within the community.

Parameter-efficient fine-tuning (PEFT) methods aim to achieve performance comparable to full-parameter fine-tuning by freezing the majority of the large model's parameters and fine-tuning a small number of parameters on downstream tasks \cite{ding2023parameter}. Based on this fundamental idea, Low-Rank Adaptation (LoRA) technology approximates model parameter updates by introducing two low-rank matrices, and it has garnered widespread attention within the community in recent years \cite{hu2022lora}. Mathematically, the original model weight $W$ can be reparametered into $W = W_0 + BA$, where $W \in \mathbb{R} ^ {m \times n} $ and $B \in \mathbb{R} ^ {m \times r}$, $A \in \mathbb{R} ^ {r \times n}$. Because of the rank $r \ll \min\{m, n\}$, the learnable parameters are far smaller than the original weight parameter count, which saves much GPU memory. Despite LoRA's high flexibility and broad applicability, its performance is constrained by the rank $r$ of the low-rank matrices $A$ and $B$, resulting in it still slightly underperforming compared to full-parameter fine-tuning \cite{xia2024chain}. 

\begin{figure*}[t]
    \centering
    \includegraphics[width=0.9\textwidth]{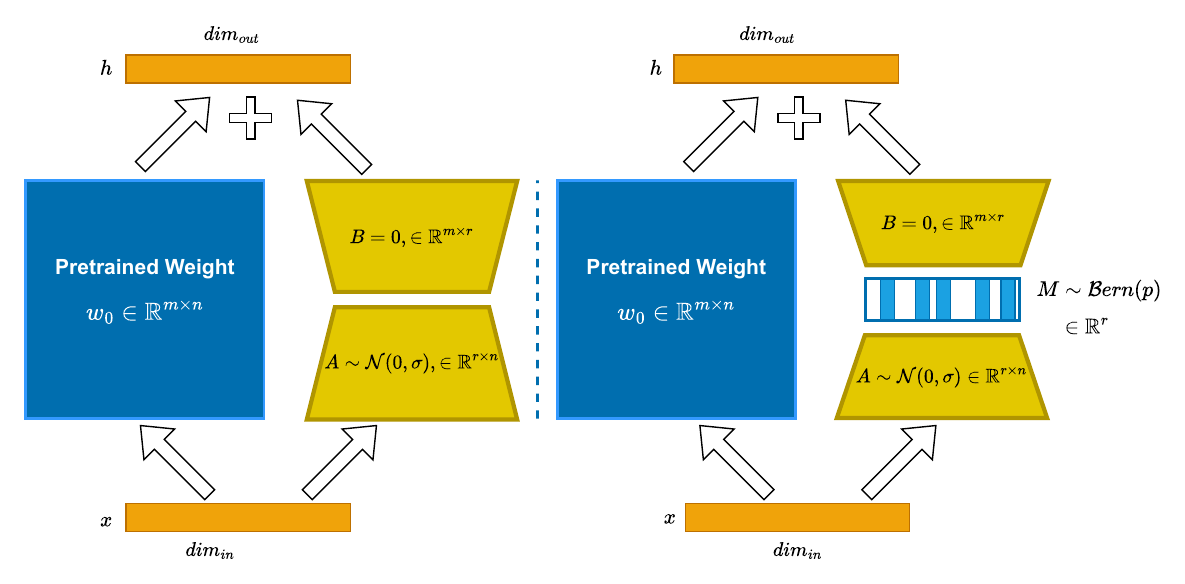} 
    \caption{Schematic comparison of LoRA (left) and DropLoRA (right). In LoRA, the original weights remain unchanged, with updates being applied solely to two low-rank matrices. DropLoRA introduces a mask matrix $M$ between these two low-rank matrices to enable pruning. At each parameter iteration step, a distinct $M$ is sampled from a Bernoulli distribution, enabling subspace learning. In both scenarios, the low-rank matrices and vectors can be seamlessly integrated into the original weight matrix $W$, thereby introducing no additional latency.}
    \label{fig:system}
\end{figure*}

To address the performance limitations of LoRA, the research community has investigated a diverse array of strategies. A number of works have focused on the initialization of LoRA, employing singular value decomposition (SVD) of the original matrices to optimize the initialization of the low-rank matrices $A$ and $B$ \cite{meng2024pissaprincipalsingularvalues, wang2025miloraharnessingminorsingular,lingam2024svftparameterefficientfinetuningsingular,buyukakyuz2024olora}. Another line of research aims to enhance the rank of LoRA by refining the low-rank matrices to mitigate the rank bottleneck and improve its expressive power \cite{meng2024periodiclora, wang2025miloraharnessingminorsingular, jiang2024morahighrankupdatingparameterefficient}. Additionally, some techniques dynamically adapt the rank of LoRA for different weights, offering greater flexibility and efficiency \cite{valipour2022dylora, zhang2023adalora}.

In contrast to reparameterizing the original weights, \citet{zhao2024galorememoryefficientllmtraining} introduces GaLore, an innovative approach that achieves memory efficiency by projecting gradients onto diverse low-rank subspaces—effectively reparameterizing the gradients—while demonstrating exceptional performance. While Galore utilizes gradient-based low-rank projection, it fundamentally differs from LoRA, representing a distinct methodology. A key distinction is that LoRA fine-tunes only a subset of parameters, whereas GaLore optimizes all parameters. Inspired by the efficiency of the dynamic subspace learning of GaLore, we are prompted to investigate: Can LoRA further enhance its performance through the application of dynamic subspace learning?

Building on this insight, we propose DropLoRA, a strategy that simulates subspace learning by dynamically adjusting the rank of LoRA. For a fixed rank, LoRA operates within a consistent low-rank subspace, with the learning subspace remaining static throughout the process. To simulate dynamic subspace learning, we propose a simple yet effective pruning strategy, as illustrated in Figure~\ref{fig:system}. By applying unified dynamic pruning to the two low-rank matrices, each pruning operation corresponds to a distinct subspace. Specifically, we sample the rank-dimension pruning matrix $M$ from a Bernoulli distribution, hence, 
$M \in \{0,1\}^r, M_i \stackrel{\text{i.i.d.}}{\sim} \mathrm{Bernoulli}\left(p\right), i \in \{1, 2, ..., r\}$, where $p$ is the pruning probability, $r$ is the rank of $A$ and $B$. Hence our DropLoRA can be formulated as $W = W_0 + (B \odot M) \times (M \odot A)$. The dynamics of subspace learning are reflected in sampling different pruning matrices $M$ at each iteration step.

Extensive experiments show that subspace learning, as exemplified by DropLoRA, can serve as a novel optimization direction. In summary, our main contributions are as follows:
\begin{itemize}
\item We propose DropLoRA, an innovative optimization strategy for LoRA, which for the first time introduces subspace learning into the LoRA framework, exploring a novel direction for optimization in the community.
\item Our pruning strategy is designed to be seamlessly integrated into any LoRA variant without introducing additional computational or storage overhead, showcasing its adaptability and practicality.
\item DropLoRA achieves state-of-the-art (SOTA) performance across diverse domains, highlighting its broad applicability and robustness.
\end{itemize}

\section{Related Work}
\textbf{Parameter-Efficient Fine-Tuning (PEFT)} methods for supervised fine-tuning of large models have become increasingly significant, particularly in resource-constrained scenarios. The development of various efficient fine-tuning methods has emerged as a prominent research focus. Existing efficient fine-tuning techniques can be categorized into the following aspects.
Methods based on adapters aim to insert different adapter layers between the layers of a model for various downstream tasks \cite{houlsby2019parameter,he2021towards, mahabadi2021parameter}.

Prompt-based methods, such as P-tuning \cite{liu2021p} and prefix-tuning \cite{li2021prefix} , introduce continuous prompt tokens into the input space, allowing for efficient adaptation of large PLMs by only fine-tuning these learned prompt embeddings while keeping the original model parameters frozen. These approaches differ from traditional fine-tuning, as they avoid direct modification of the underlying model weights, instead relying on task-specific soft prompts to guide the model's behavior. However, both adapter-based and prompt-based approaches modify the model's internal structure, either by inserting additional trainable layers or by prepending learnable prompt embeddings. While these methods significantly reduce the number of trainable parameters compared to full fine-tuning, they inevitably introduce additional computational overhead during both training and inference. Specifically, the inclusion of extra parameters or prompt tokens increases memory usage and may lead to higher inference latency, particularly in real-time applications where low-latency responses are critical.  

LoRA-based methods and their variants achieve parameter efficiency by decomposing the base weight matrix into two low-rank matrices, demonstrating significant advantages in deployment, particularly for mobile device applications \cite{hu2022lora, kopiczko2023vera, meng2024periodiclora, liu2024dora, zhang2023adalora, meng2024pissaprincipalsingularvalues}. For diverse applications, it is only necessary to store distinct LoRA adapters specifically fine-tuned for their respective downstream tasks. Variants of LoRA primarily include optimization of initialization parameters, rank enhancement, and adaptive rank selection, among others \cite{meng2024pissaprincipalsingularvalues, meng2024periodiclora, valipour2022dylora}. Extensive research related to LoRA demonstrates that LoRA-based methods are currently dominating the field of Parameter-Efficient Fine-Tuning (PEFT).

\textbf{Subspace Learning} focuses on deriving low-dimensional, essential features from high-dimensional data to enhance learning efficiency and effectiveness. By eliminating redundancy and capturing essential characteristics, subspace learning facilitates more efficient and effective learning processes, enhancing both computational performance and model accuracy \cite{liu2012robust, de2003framework}. Extensive research has demonstrated that subspace learning exhibits excellent generalization capabilities, making it a robust approach for various machine learning tasks \cite{hinton2006reducing, wright2008robust, zhao2024galorememoryefficientllmtraining}. LoRA assumes that weight updates occur in a low-rank space; however, due to its static rank nature, it can essentially be regarded as a form of static subspace learning.

\begin{algorithm}[t]
   \caption{DropLoRA, torch-style pseudocode.}
   \label{algo:DropLoRA}
   \definecolor{codeblue}{rgb}{0.25,0.5,0.5}
    \lstset{
      basicstyle=\fontsize{7.8pt}{7.8pt}\ttfamily\bfseries,
      commentstyle=\fontsize{7.8pt}{7.8pt}\color{codeblue},
      keywordstyle=\fontsize{7.8pt}{7.8pt},
    }
\begin{lstlisting}[language=python]
class DropLoRALayer(nn.Module):
    def __init__(
    self,
    r: int = 32, # rank
    p: float = 0.5, # pruning probability
    d1: int = 4096, # input dimension
    d2: int = 4096, # output dimension
    base_layer: nn.Module # pre-trained layer
    ):
    self.base_layer = base_layer
    self.A = torch.randn(r, d1)
    self.B = torch.zeros(d2, r)
    self.M = Dropout(p) ## Line 1.
    self.base_layer.freeze()

    def forward(self, x: torch.Tensor):
        h = self.base_layer(x)
        ## In LoRA
        ## delta = x @ self.A @ self.B
        ## Line 2
        delta = self.M(x @ self.A) @ self.B  
        
        return h + delta
\end{lstlisting}
\end{algorithm}

\section{Method}
LoRA reparameterizes the update of the original weights as the product of two low-rank matrices, as expressed in Equation~\ref{eq:lora}:

\begin{equation}
    h = W_0x + \underline{BA}x
\label{eq:lora}
\end{equation}

where $W_0 \in \mathbb{R} ^ {m \times n} $ and $B \in \mathbb{R} ^ {m \times r}$, $A \in \mathbb{R} ^ {r \times n}$. During the training process, the original weights $W_0$ remain unchanged, with updates being applied exclusively to the weights of the two low-rank matrices $A, B$. Here, We undeline the parameters updated during the training process. Because of the rank $r \ll \min\{m, n\}$, the learnable parameters are far smaller than the original weight parameter count. When the rank is fixed, LoRA can be viewed as learning within a static subspace, which may inherently constrain its expressive capacity. 

To simulate dynamic subspace learning, we propose a pruning technique based on dynamic masking. Specifically, we sample the rank dimension using a Bernoulli distribution $\mathrm{Bernoulli}\left(p\right)$ to generate a mask vector of rank size, where a value of $1$ retains the dimension and a value of $0$ discards it. Our DropLoRA method is expressed as:
\begin{equation}
    h = W_0x + \left(\underline{B} \odot M\right) \left(M \odot \underline{A}\right)x
\label{eq:droplora}
\end{equation}
where $M \in \{0,1\}^r, M_i \stackrel{\text{i.i.d.}}{\sim} \mathrm{Bernoulli}\left(p\right), i \in \{1, 2, ..., r\}$ and $\odot$ represents element-wise multiplication. $M$ is a mask matrix, where $p$ represents the pruning probability.
At each training iteration step, we randomly sample a distinct mask matrix $M$ to simulate dynamic subspace learning. 

\paragraph{Rank Analysis}When we apply the sampled mask to prune the rank dimension, it implies that the effective rank of the two low-rank matrices $\tilde{A} = \left( M \odot A \right)$ and $\tilde{B} = \left(B \odot M \right)$ is reduced compared to their original rank. With a pruning probability of $0.5$, the rank of $\tilde{A}$ and $\tilde{B}$ becomes only half of the original rank.

\paragraph{Equivalence} Intuitively, there are two pruning strategies: one applies a unified pruning using the same mask matrix for both low-rank matrices, while the other prunes $A$ and $B$ separately using distinct mask matrices. For the latter, the two mask matrices operate under a logical AND relationship, effectively equivalent to their intersection. This is functionally identical to using a single mask matrix with values equal to the intersection. Thus, the two approaches are equivalent.

\paragraph{Easy Implementation}
Since the product of LoRA’s two low-rank matrices is mathematically equivalent to a two-layer perceptron without activation, the masked pruning strategy effectively functions as dropout applied to the intermediate hidden layer. This implies that, compared to LoRA, DropLoRA can be implemented with just two additional lines of code, as illustrated in Algorithm~\ref{algo:DropLoRA}.

It is worth noting that, although similar in implementation, our method differs from traditional Dropout regularization methods \cite{srivastava2014dropout}. The Dropout method generally randomly drops some of the high-dimensional inputs. In contrast, our method randomly discards the rank dimension of LoRA low-rank matrices. Intuitively, the expressive power of LoRA is limited by the size of the rank. Randomly discarding the rank dimension will further reduce the expressive power of LoRA, resulting in severe performance degradation. 
Therefore, pruning the rank dimension is somewhat counterintuitive.
However, dynamic low-rank subspace learning allows DropLoRA to
overcome the limitations of traditional LoRA,
which operates within a static subspace and the model is prompted to learn more intrinsic parameter variation characteristics, thereby improving performance.

\paragraph{Training and Inference} During the training process, at each iteration step, we obtain different low-rank subspaces by sampling different pruning vectors through the Bernoulli distribution. During backpropagation, only the retained parameters are involved in the update. In the reasoning process, in order to enhance the model's expressive power, we do not use the pruning module. By integrating the parameters learned in different subspaces, this has a similar effect to ensemble learning, thereby improving the robustness of the model.

\begin{table*}[ht]
\centering
\resizebox{1.0\textwidth}{!}
{
\begin{tabular}{lccccccccccc}
\toprule
\textbf{Model}& \textbf{PEFT} &\textbf{\# Parameters} & \textbf{BoolQ} & \textbf{PIQA} & \textbf{SIQA} & \textbf{HellaSwag} & \textbf{WinoGrande} & \textbf{ARC-e} & \textbf{ARC-c} & \textbf{OBQA} & \textbf{Avg.} \\
\midrule
ChatGPT$^\dagger$ & $-$ & $-$ & 73.1 & 85.4 & 68.5 & 78.5 & 66.1 & 89.8 & 79.9 & 74.8 & 77.0 \\

\midrule
\multirow{9}{*}{LLaMA2-7B} & LoRA$^\dagger$ & 56.10M & 69.8 & 79.9 & 79.5 & 83.6 & 82.6 & 79.8 & 64.7 & 81.0 & 77.6 \\
& DoRA$^\dagger$ & 56.98M  & 71.8 & 83.7 & 76.0 & 89.1 & 82.6 & 83.7 &
68.2 & 82.4 & 79.7 \\
& PiSSA$^\star$ & 56.10M & 67.6 & 78.1 & 78.4 & 76.6 & 78.0 & 75.8 & 60.2 & 75.6 & 73.8 \\
& MiLoRA$^\dagger$ & 56.10M & 67.6 & 83.8 & 80.1 & 88.2 & 82.0 & 82.8 & 68.8 & 80.6 & 79.2 \\
\cline{2-12}
& LoRA & 56.10M & 74.37 & \textbf{87.38} & \textbf{81.32} & 95.07 & 85.79 & 88.80 & 75.34 & \underline{87.00} & 84.38 \\
& DoRA & 56.98M  & \underline{74.46} & 86.18 & 80.60 & 94.91 & \underline{87.53} & 89.14 &
75.85 & 86.40 & 84.39 \\
& PiSSA & 56.10M & 73.27 & 82.59 & 79.84 & 92.88 & 84.77 & 85.23 & 71.33 & 84.80 & 81.84 \\
& MiLoRA & 56.10M & \textbf{74.53} & 86.45 & 80.81 & \underline{95.23} & 86.90 & \underline{89.48} & \underline{76.54} & 86.00 & \underline{84.49} \\
& DropLoRA (Ours) & 56.10M & 74.22 & \underline{87.00} & \underline{80.91} & \textbf{95.24} & \textbf{87.61} & \textbf{89.73} & \textbf{77.30} & \textbf{87.20} & \textbf{84.91} \\

\midrule
\multirow{9}{*}{LLaMA3-8B} & LoRA$^\dagger$ & 56.62M & 70.8 & 85.2 & 79.9 & 91.7 & 84.3 & 84.2 & 71.2 & 79.0 & 80.8 \\
& DoRA$^\dagger$ & 57.41M  & 74.6 & 89.3 & 79.9 & 95.5 & 85.6 & 90.5 &
80.4 & 85.8 & 85.2 \\
& PiSSA$^\star$ & 56.62M & 67.1 & 81.1 & 77.2 & 83.6 & 78.9 & 77.7 & 63.2 & 74.6 & 75.4 \\
& MiLoRA$^\dagger$ & 56.62M & 68.8 & 86.7 & 77.2 & 92.9 & 85.6 & 89.48 & 75.5 & 81.8 & 81.9 \\
\cline{2-12}
& LoRA & 56.62M & 75.54 & 89.06 & 81.12 & 95.99 & 88.08 & 92.80 & 82.59 & 89.20 & 86.78 \\
& DoRA & 57.41M  & 75.41 & 89.12 & 881.27 & 95.83 & 87.69 & 92.42 &
82.59 & 89.00 & 86.67 \\
& PiSSA & 56.62M & 72.66 & 86.13 & 80.14 & 93.60 & 84.77 & 89.73 & 76.88 & 86.00 & 83.74 \\
& MiLoRA & 56.62M & 74.53 & 86.45 & 80.81 & 95.23 & 86.90 & 89.48 & 76.54 & 86.00 & 84.49 \\
& DropLoRA (Ours) & 56.62M & \textbf{76.45} & \textbf{90.04} & \textbf{82.19} & \textbf{96.59} & \textbf{89.34} & \textbf{93.18} & \textbf{83.28} & \textbf{89.80} & \textbf{87.61} \\
\bottomrule
\end{tabular}
}
\caption{Commonsense reasoning evaluation results for LLaMA2-7B and LLaMA3-8B on eight tasks. The reported metric in this table is accuracy. $^\dagger$Results are cited from the original paper and $^\star$results are cited from \citet{wang2025miloraharnessingminorsingular}. For PEFT results, all other experiments without superscripts are performed by ourselves. Bold numbers indicate the highest performance scores and underline numbers indicate the second performance scores for each dataset across the different PEFT methods for the corresponding model.}
\label{tab:commonsense}
\end{table*}

\section{Experiments}
To evaluate the effectiveness of the DropLoRA method, we conducted extensive experiments encompassing commonsense reasoning tasks, mathematical tasks, coding tasks, instruction following tasks. For all tasks, we choose the same LoRA-related baselines including:\\
\textbf{LoRA}\cite{hu2022lora} decomposes a parameter update into the product of two low-rank matrices, where one matrix is initialized with Gaussian distribution and the other is initialized with zeros.\\
\textbf{DoRA}\cite{liu2024dora} decouples the magnitude and direction of the parameter update, using LoRA to update the direction and a learnable magnitude vector to update the magnitude.\\
\textbf{PiSSA}\cite{meng2024pissaprincipalsingularvalues} initializes LoRA by applying singular value decomposition (SVD) to the pre-trained weights, using the \textbf{Principal Singular Components} to initialize LoRA, while the residual components are used to initialize the pre-trained weights.\\
\textbf{MiLoRA}\cite{wang2025miloraharnessingminorsingular} initializes LoRA by applying singular value decomposition (SVD) to the pre-trained weights, using the \textbf{Minor Singular Components} to initialize LoRA, while the residual components are used to initialize the pre-trained weights.\\
All experiments are conducted on $4 \times A100$ GPUs with Deepspeed ZERO-2 stage\cite{rasley2020deepspeed} to accelerate training.

\subsection{Commensense Reasoning}
To evaluate the impact of efficient fine-tuning techniques on commonsense knowledge and logical reasoning abilities, we conduct experiments on commonsense knowledge reasoning tasks.

\paragraph{Datasets} The commonsense reasoning dataset consists of 8 sub-tasks, each with its own predefined training and testing sets, including BoolQ\cite{clark2019boolq}, PIQA\cite{bisk2020piqa}, SIQA\cite{sap2019socialiqa}, HellaSwag\cite{zellers2019hellaswag}, WinoGrande\cite{sakaguchi2021winogrande}, ARC-e, ARC-c\cite{clark2018think} and OBQA\cite{mihaylov2018can}. We follow the experimental setup from \citet{hu2023llm}, where the training sets of the 8 sub-tasks are combined, and inference and evaluation are conducted separately on their respective testing sets. 

\paragraph{Experimental Setting}  We choose LLaMA2-7B\cite{touvron2023llama}\footnote{\url{https://hf-mirror.com/meta-llama/Llama-2-7b-hf}\label{llama2}} and LLaMA3-8B\cite{grattafiori2024llama}\footnote{\url{https://hf-mirror.com/meta-llama/Meta-Llama-3-8B}\label{llama3}} as our backbone models. To ensure a fair comparison, we implement all PEFT experiments ourselves. We also report chatGPT-api based results sourced from \citet{liu2024dora}. For the hyperparameter configuration, we also follow the parameter settings from \citet{hu2023llm}.  Note that, to accelerate training, we use a batch size of $128$ instead of the original configuration of $16$. For all other hyperparameters, we strictly follow the parameter settings from \citet{hu2023llm}. It is important to note that in all experiments, for DropLoRA, we only adjust the pruning probability and do not adjust any other hyperparameters. For detailed hyperparameter configurations, see Appendix \ref{sec:appendix}.

\paragraph{Result} Table~\ref{tab:commonsense} 
presents the experimental results for the common-sense reasoning task. We also report the evaluation results based on the ChatGPT API as outlined in the DoRA paper \cite{liu2024dora}, which are obtained with the GPT-3.5-turbo API using a zero-shot Chain of Thought approach. 

As can be seen, on the LLaMA2-7B, DropLoRA achieved the best performance on five datasets (HellaSwag, WinoGrande, ARC-e, ARC-c, OBQA), the second-best performance on two datasets (PIQA, SIQA), and the best average performance across all eight datasets with an average performance increase of $+0.53$ points compared to 
LoRA. On the LLaMA3-8B model, DropLoRA achieves the best performance on all eight datasets, with an average performance increase of $+0.83$ points compared to LoRA, indicating that DropLoRA is an effective parameter-efficient fine-tuning method.
We observe that both LoRA and DoRA achieve comparable performance on LLaMA2-7B and LLaMA3-8B. MiLoRA slightly outperforms LoRA on LLaMA2-7B, but shows a significant performance gap on LLaMA3-8B. PiSSA, on the other hand, performs substantially worse than other methods on both models. This indicates instability in performance for methods that fine-tune either the principal or the minor singular components. In contrast, our method achieves the best performance on both models, demonstrating its superior stability.

\begin{table}[t]
\centering
\resizebox{0.48\textwidth}{!}
{
\begin{tabular}{cccccc}
\toprule
\textbf{Model} &\textbf{Method} & \textbf{\# Parameters} & \textbf{GSM8K} & \textbf{MATH} & \textbf{Average} \\
\midrule

\multirow{9}{*}{LLaMA2-7B} 
&Full FT$^\dagger$ & 6738M  & 66.5 & 19.8 & 43.2 \\ 
&LoRA$^\dagger$ & 112.20M & 60.6 & 16.9 & 38.7 \\
&PiSSA$^\dagger$ & 112.20M & 58.2 & 15.8 & 37.0 \\
&MiLoRA$^\dagger$ & 112.20M & 63.5 & 17.8 & 40.7 \\

\cline{2-6}
&LoRA & 112.20M & 65.66 & 16.02 & 40.84 \\
&DoRA & 113.07M & 66.19 & 16.14 & 41.16  \\
&PiSSA & 112.20M & 64.37 & 15.96 & 40.16 \\
&MiLoRA & 112.20M & 64.52 & 14.92 & 39.72 \\
&DropLoRA (Ours) & 112.20M & \textbf{66.72} & \textbf{16.38} & \textbf{41.55} \\

\midrule
\multirow{5}{*}{LLaMA3-8B} 
&LoRA & 113.25M & 80.44 & 30.46 & 55.45 \\
&DoRA & 114.03M & 80.44 & 30.21 & 55.32  \\
&PiSSA & 113.25M & 79.53 & 28.92 & 54.22 \\
&MiLoRA & 113.25M & 80.74 & 30.62 & 55.68 \\
&DropLoRA (Ours) & 113.25M & \textbf{81.32} & \textbf{30.74} & \textbf{56.03} \\
\bottomrule
\end{tabular}
}
\caption{Math reasoning evaluation results for GSM8K and MATH based on LLaMA2-7B and LLaMA3-8B. $^\dagger$Results are cited from \citet{wang2025miloraharnessingminorsingular} and All other experiments without superscripts are performed by ourselves.}
\label{tab:math}
\end{table}

\subsection{Math and Code Reasoning} 
To evaluate numerical computation and logical reasoning capabilities, we conduct performance assessments on mathematical problem-solving and programming tasks.

\paragraph{Datasets}We evaluate mathematical problem-solving capabilities on MetaMathQA dataset\cite{yu2023metamath}, including $395$K samples generated by augmenting the training sets of GSM8K\cite{cobbe2021training} and MATH\cite{hendrycks2021measuring}. During the testing phase, we perform inference and evaluate performance on the test sets of GSM8K and MATH separately.

To evaluate code capabilities, we fine-tune on the CodeFeedback\cite{zheng2024opencodeinterpreter} dataset and perform evaluation on the HumanEval\cite{chen2021evaluating}and MBPP\cite{austin2021program} test sets.

\begin{table}[t]
\centering
\resizebox{0.48\textwidth}{!}
{
\begin{tabular}{cccccc}
\toprule
\textbf{Model}&\textbf{Method}& \textbf{\# Parameters} &\textbf{HumanEval} & \textbf{MBPP} & \textbf{Average} \\
\midrule
LLaMA2-7B &Full FT$^\dagger$ & 6738M & 21.34 & 35.59 & 28.47 \\ \midrule
\multirow{5}{*}{LLaMA2-7B} 
&LoRA & 56.10M & 18.90 & 41.27 & 30.09 \\
&DoRA & 56.98M & 14.63 & 42.86 & 28.75  \\
&PiSSA & 56.10M & 9.76 & 42.86 & 26.31 \\
&MiLoRA & 56.10M & \textbf{21.34} & 37.57 & 29.46 \\
&DropLoRA (Ours) & 56.10M & \textbf{21.34} & \textbf{43.39} & \textbf{32.37} \\

\midrule
\multirow{5}{*}{LLaMA3-8B} 
&LoRA & 56.62M & \textbf{62.81} & 65.87 & 64.34 \\
&DoRA & 57.41M & 59.76 & 66.40 & 63.08  \\
&PiSSA & 56.62M & 57.93 & 65.87 & 61.90 \\
&MiLoRA & 56.62M & 59.76 & 66.93 & 63.35 \\
&DropLoRA (Ours) & 56.62M & 60.37 & \textbf{70.37} & \textbf{65.37} \\
\bottomrule
\end{tabular}
}
\caption{Code evaluation results for HumanEval and MBPP based on LLaMA2-7B and LLaMA3-8B. $^\dagger$Results are cited from \citet{meng2024pissaprincipalsingularvalues} and the other experimental results are from ourselves.}
\label{tab:code}
\end{table}

\paragraph{Experimental Setting} We choose LLaMA2-7B\footref{llama2} and LLaMA3-8B\footref{llama3} as our pre-trained models. We use hyperparameter configurations similar to those for commonsense reasoning. For the mathematical reasoning task, due to the large training set of MetaMathQA, we set the rank of LoRA to $64$ and train for only one epoch to avoid overfitting. For the code evaluation task, we maintain the same hyperparameter configuration as for commonsense reasoning. For detailed hyperparameter configurations, see Appendix \ref{sec:appendix}.

\paragraph{Result} 
Tables~\ref{tab:math} and Table~\ref{tab:code} present the experimental results for mathematical reasoning and code reasoning tasks, respectively. DropLoRA consistently achieves state-of-the-art performance across all four reasoning tasks, demonstrating its effectiveness in handling both mathematical and code-based problem-solving scenarios. 

Notably, on mathematical reasoning tasks with LLaMA2-7B, DropLoRA outperforms standard LoRA by an average margin of $+0.7$ percentage points, while this advantage expands to $+2.28$ percentage points on coding tasks. The performance gap persists with LLaMA3-8B, where DropLoRA achieves $+0.58$ and $+1.03$ percentage point improvements over LoRA in mathematical and coding tasks, respectively.

We also observed that LoRA and DoRA achieved comparable performance on mathematical reasoning tasks, while MiLoRA exhibited significant performance fluctuations across two models. On coding tasks, DoRA, MiLoRA, and PiSSA all show substantial performance gaps compared to LoRA, hinting at the complexity of coding tasks. Despite this, our method still significantly outperformed LoRA. Specifically, on LLaMA2-7B, it surpassed LoRA by $+2.3$ percentage points; on LLaMA3-8B, it surpassed LoRA by $+1$ point. The absolute leading advantage demonstrates the superior performance of our method on reasoning tasks.

\begin{table}[t]
\centering
\resizebox{0.48\textwidth}{!}
{
\begin{tabular}{lccc}
\toprule
\textbf{Model}&\textbf{Method}& \textbf{\# Parameters} & \textbf{MT-Bench} \\
\midrule
\multirow{5}{*}{LLaMA2-7B} 
&LoRA & 56.10M & 5.16  \\
&DoRA & 56.98M  & 5.33  \\
&PiSSA & 56.10M  & 5.20  \\
&MiLoRA & 56.10M & 5.25  \\
&DropLoRA (Ours) & 56.10M & \textbf{5.54} \\

\midrule
\multirow{5}{*}{LLaMA3-8B} 
&LoRA & 56.62M & 6.31  \\
&DoRA & 57.41M  & 6.09  \\
&PiSSA & 56.62M  & 5.94  \\
&MiLoRA & 56.62M & 6.11  \\
&DropLoRA (Ours) & 56.62M & \textbf{6.42} \\
\bottomrule
\end{tabular}
}
\caption{Instruction following results based on LLaMA2-7B and LLaMA3-8B, assigned by GPT-4 to the answers. All experimental results are conducted by ourselves.}
\label{tab:IF}
\end{table}

\begin{figure*}[t]
    \centering
    \includegraphics[width=0.85\textwidth]{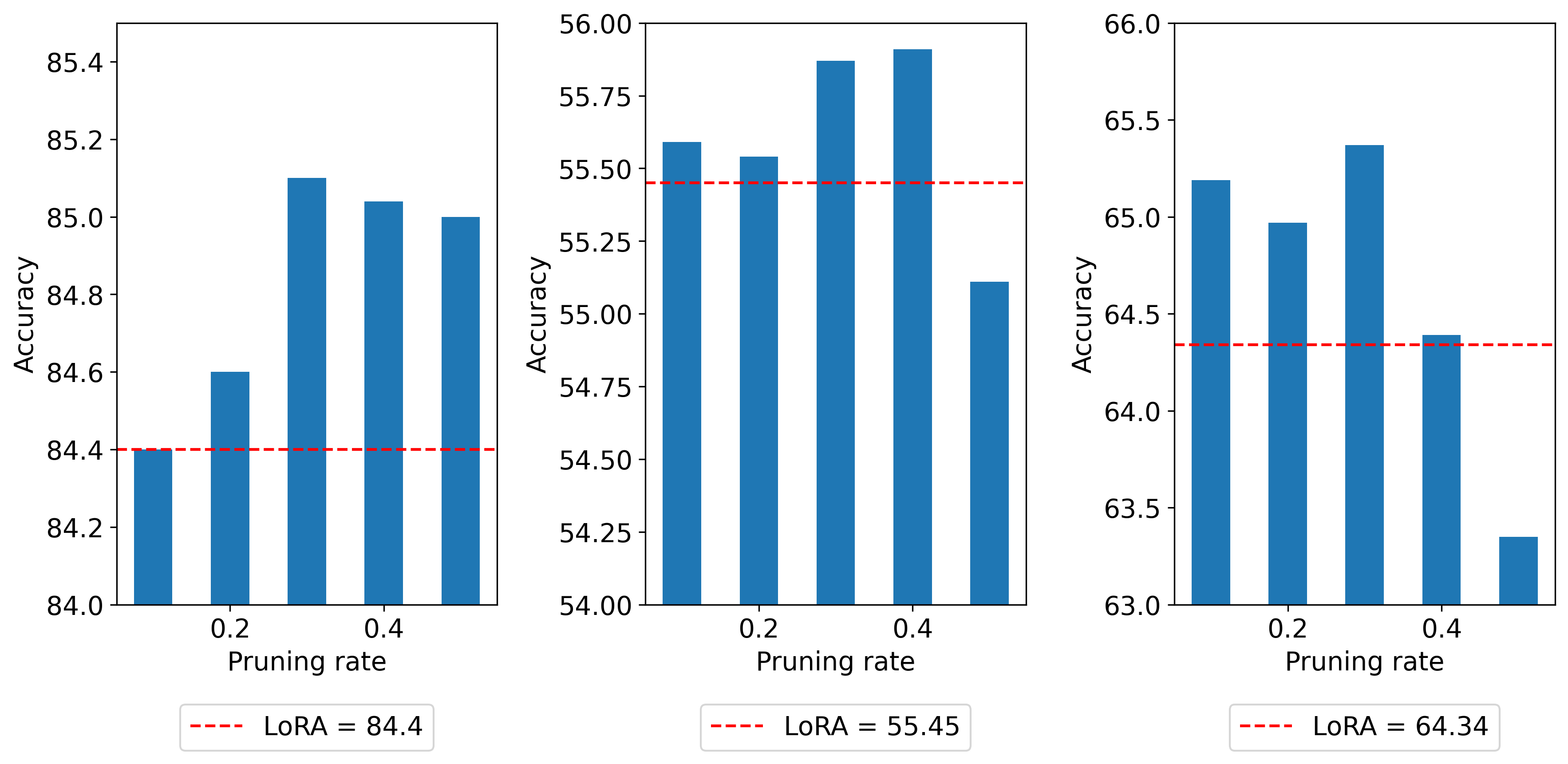} 
    \caption{Average accuracy of LoRA and DropLoRA for varying pruning rate
     on the commonsense reasoning, math and code tasks. The left, middle, and right figures correspond to commonsense reasoning, math, and coding tasks respectively. }
    \label{fig:pruning rate}
\end{figure*}

\subsection{LLM Capability for Open Questions}

To comprehensively evaluate our model's capacity for handling open-ended questions and executing complex instructions, we employ the MT-Bench dataset \cite{zheng2023judging}, a widely recognized benchmark in the field of natural language processing. This meticulously curated dataset contains 80 carefully designed questions spanning diverse domains and difficulty levels, along with 3,300 expert-annotated pairwise human preference judgments comparing responses generated by six different models. 

\paragraph{Experimental Setting} We utilize the same hyperparameters as those used in the \citet{hu2023llm}. For the evaluation of conversational abilities, we employ the method mentioned in the MT-Bench paper\cite{zheng2023judging}\footnote{\url{https://github.com/lm-sys/fastchat}}, utilizing GPT-4 to score the dialogue tasks. For detailed hyperparameter configurations, see Appendix \ref{sec:appendix}.

\paragraph{Result} Table~\ref{tab:IF} displays the results of our experiments conducted on the dialogue task. Our proposed method demonstrates the best performance across both models. Specifically, when compared to LoRA, the performance improves by +0.38 points on LLaMA2-7B and by +0.38 points on LLaMA3-8B. Notably, we observe that both PiSSA and MiLoRA, in comparison to LoRA, yield nearly the same marginal gains in the dialogue task. This suggests that the differences in performance between fine-tuning the principal singular component (PiSSA) and fine-tuning the minor singular component (MiLoRA) are relatively minor and do not have a significant impact on this particular task.

\begin{table}[t]
\centering
\resizebox{0.45\textwidth}{!}
{
\begin{tabular}{c|ccc}
\hline
Method & rank & pruning rate &accuracy \\
\hline
LoRA &16 & 0 & 84.5 \\
LoRA &32 & 0 & 84.4 \\
DropLoRA (Ours) &32 &0.5 &\textbf{84.9}\\
\hline
\end{tabular}
}
\caption{The performance comparison of LoRA and DropLoRA on inference tasks with different ranks and pruning rates. }
\label{tab:cmp}
\end{table}

\subsection{Study} 
\paragraph{Effect of Pruning Module} 
DropLoRA inserts a pruning module between the two low-rank matrices of LoRA to simulate subspace learning, while keeping everything else consistent with LoRA. As can be seen from Table~\ref{tab:commonsense} $\sim$ Table~\ref{tab:IF}, on all four tasks, whether it is on the LLaMA2-7B or LLaMA3-8B model, the performance of DropLoRA is significantly better than that of LoRA, proving the effectiveness of the pruning module and its generalization to different tasks and models. As shown in Table ~\ref{tab:cmp}, for DropLoRA, when the rank is 32 and the pruning rate is 0.5, it means that only half of the parameters are updated during each parameter update, which is comparable to the LoRA parameters with rank 16. However, regardless of whether the rank of LoRA is 16 or 32, DropLoRA consistently outperforms LoRA, proving the effectiveness of the DropLoRA pruning module.

\paragraph{Effect of Pruning Rate} 
We explore the impact of different pruning rates on experimental results, such as the pruning rate in Equation~\ref{eq:droplora}.
Figure~\ref{fig:pruning rate} shows the fine-tuning performance of different pruning rates on the commonsense reasoning, math and coding tasks. We can see that when the pruning rate varies within the range of $0.1 \sim 0.5$, the performance fluctuates slightly. When the pruning rate is $0.3$, compared with LoRA, the performance improvement is the greatest. We observe that when the pruning rate is set to $0.5$, it starts to perform worse than LoRA on math and code tasks. This is because when the pruning rate is too large, it will reduce the low-rank subspace representation ability of the model, resulting in performance degradation. We also observe that even when the pruning rate is set to $0.5$, which means that only half of the parameters are activated during the training process of each subspace, DropLoRA can still achieve better performance than LoRA on the commonsense reasoning task. This demonstrates the effectiveness of subspace learning.

\begin{figure}[t]
    \centering
    \includegraphics[width=0.45\textwidth]{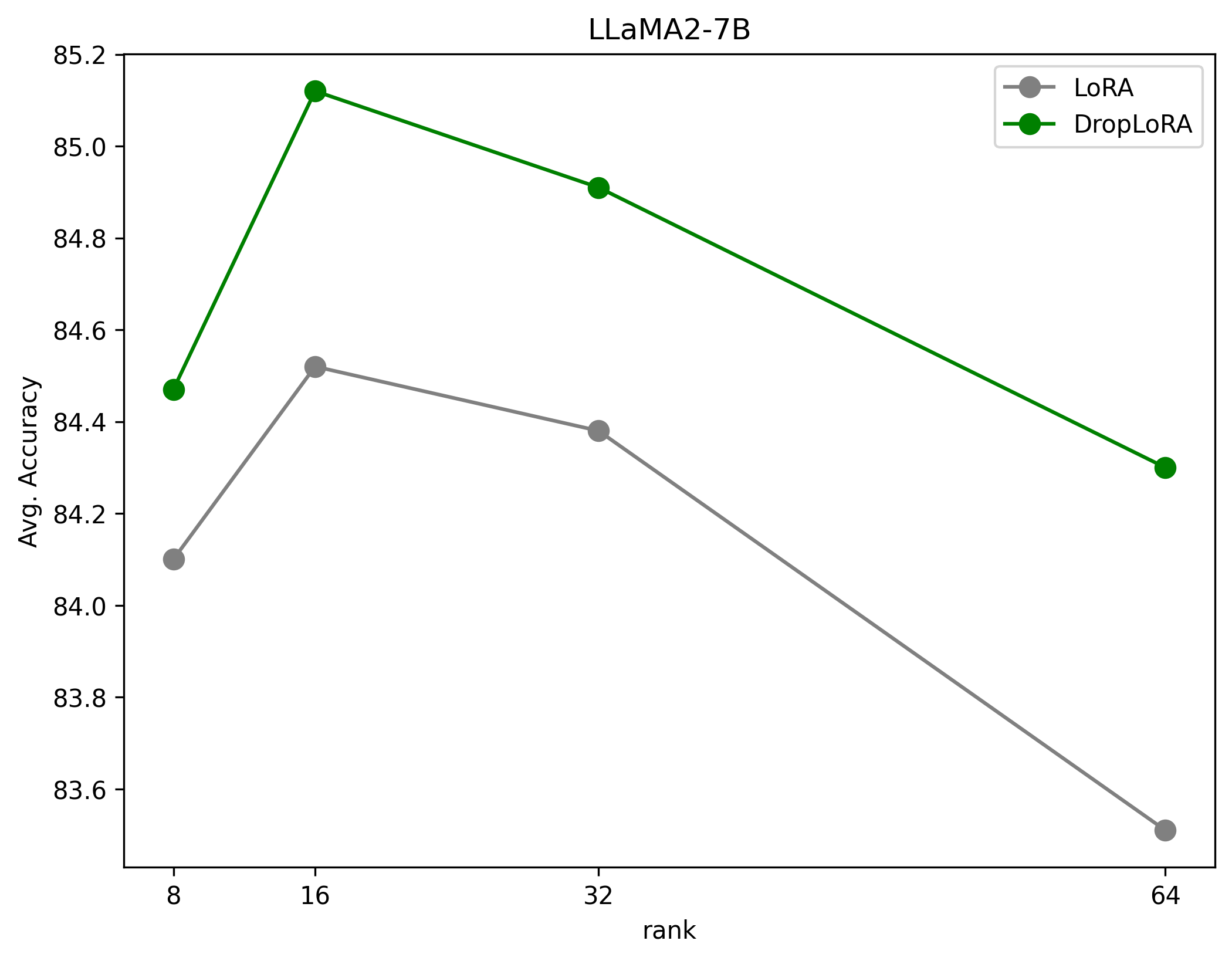} 
    \caption{Average accuracy of LoRA and DropLoRA for varying ranks
     for LLaMA-7B on the commonsense reasoning tasks.}
    \label{fig:rank}
\end{figure}

\paragraph{Parameter Scalability} We conduct an exploration into the relationship that exists between the quantity of trainable parameters and the performance of both the Low-Rank Adaptation (LoRA) method and our proposed method. We set the rank  $r = \{8, 16, 32, 64\}$, and $\alpha$ remains twice the rank. Other hyperparameters remain consistent with those of the commensense reasoning task. The average accuracy of LoRA and DropLoRA for varying ranks for LLaMA-7B on the commonsense reasoning tasks is depicted in Figure~\ref{fig:rank}. As shown in Figure~\ref{fig:rank}, under all rank configurations, DropLoRA consistently outperforms LoRA. Due to the structural similarity between the two, their performance trends are also similar. When the rank is larger, DropLoRA's performance remains significantly superior to that of LoRA. However, when the rank is smaller, the performance gap between the two narrows. This is because, when the rank is small, DropLoRA, due to the pruning module, learns in a lower-rank subspace compared to LoRA. An excessively low rank can limit the expressive power of the subspace learning.

\section{Conclusion}
In this paper, we introduce DropLoRA, a simple yet effective low-rank adaptive method for parameter-efficient fine-tuning of large language models. By inserting a pruning module between the two low-rank matrices of LoRA to simulate subspace learning, we show that performance can be improved not only by increasing LoRA’s rank but also by lowering it. We validate the effectiveness of DropLoRA on a wide range of large language model evaluation benchmarks, including commonsense reasoning, math reasoning, code generation, and instruction-following tasks. Experimental results indicate that DropLoRA consistently outperforms other baseline methods, including LoRA, DoRA, PiSSA, and MiLoRA, across all tasks. Compared to LoRA, DropLoRA introduces no additional parameters, thus not increasing any training or inference costs. Our research shows that, in addition to increasing the rank of LoRA, lowering its rank can also enhance the performance, providing a new perspective for future optimization on parameter-efficient fine-tuning of LLMs.

\section*{Limitations}
Due to computational resource constraints, we have only validated the effectiveness of DropLoRA on large model generation tasks, such as commonsense reasoning, math reasoning, code generation, and instruction-following tasks. However, an interesting future direction is whether DropLoRA can enhance performance on multimodal large model benchmark tasks beyond language generation. Another open question is whether we can provide a theoretical foundation to support the effectiveness of rank reduction for simulating subspace learning. We consider these unresolved issues as important areas for future research.

\bibliography{custom}

\newpage
\appendix

\section{Appendix}
\label{sec:appendix}
Table~\ref{tab:data statistics.} presents the statistics of the datasets used in this paper.

\subsection{Dataset Statistics}
\begin{table}[h]
\centering
\resizebox{0.45\textwidth}{!}
{
\begin{tabular}{ccccc}
\hline
Dataset & Domain & \# train & \# test  & Answer \\
\hline
BoolQ & CS & 9.4K & 3,270 & Yes/No \\
PIQA & CS & 16.1K & 1,830 & Option \\
SIQA & CS & 33.4K & 1,954 & Option \\
HellaSwag & CS & 39.9K & 10,042 & Option \\
WinoGrande & CS &63.2K &1,267 & Option\\
ARC-e &CS &1.1K &2,376 &Option\\
ARC-c &CS &2.3K &1,172 &Option \\
OBQA  &CS &5.0K &500 & Option\\
GSM8K &Math &240K &1,319 & Number \\
MATH  &Math &155K &5,000 & Number \\
Python &Code &104,848 & 563 & Code \\
Instruction Following &Conversation &143K &80 &Text \\
\hline
\end{tabular}
}
\caption{Details of datasets used in our experiment setting including commonsense reasoning, math reasoning, code reasoning and instruction following tasks.}
\label{tab:data statistics.}
\end{table}

\subsection{Our Hyperparameter Setup for LLM}
Table~\ref{tab:settings for NLG.} presents the hyperparameter configurations used in our experiments. To ensure fairness, our hyperparameter settings are consistent with those reported in the DoRA\cite{liu2024dora} and MiLoRA\cite{wang2025miloraharnessingminorsingular} papers. Note that, to accelerate training, the batch size for all experiments in this paper is set to 128.

\begin{table}[h]
\centering
\resizebox{0.45\textwidth}{!}
{
\begin{tabular}{ccccc}
\hline
Hyperparameters & Commonsense & Math & Code  &Conversation \\
\hline
Rank $r$ & 32 & 64 & 32 & 32 \\
$\alpha$ of LoRA & 64 & 128 & 64 & 64 \\
$\alpha$ of DoRA & 64 & 128 & 64 & 64 \\
$\alpha$ of DropLoRA & 64 & 128 & 64 & 64 \\
$\alpha$ of PiSSA &32 &64 &32 &32\\
$\alpha$ of MiLoRA &32 &64 &32 &32\\
Dropout &\multicolumn{4}{c}{0.05} \\
Pruning Rate &\multicolumn{4}{c}{0.1 $\sim$ 0.5}\\
Optimizer &\multicolumn{4}{c}{AdamW}\\
LR &\multicolumn{4}{c}{3e-4} \\
LR Scheduler &\multicolumn{4}{c}{Linear}\\
Batch size &\multicolumn{4}{c}{128} \\
Warmup Steps &\multicolumn{4}{c}{100} \\
Epochs &3 &1 &3 &3 \\
Reparameterization &\multicolumn{4}{c}{Q,K,V,Up,Down}\\
\hline
\end{tabular}
}
\caption{Our hyperparameter configuration for LLM generation benchmarks for fine-tuning LLaMA2-7B, LLaMA3-8B on the commonsense reasoning, math reasoning, code reasoning and instruction following tasks.}
\label{tab:settings for NLG.}
\end{table}

\subsection{Case Study}
To provide an intuitive demonstration of the effects, we randomly sampled two cases from the mathematical reasoning tasks and presented the reasoning analysis results. Table~\ref{tab:case_study} and Table~\ref{tab:case_study1} show the inference results of various methods. In Table~\ref{tab:case_study}, except for DoRA, the reasoning processes and outcomes of the other methods are correct. Although DoRA produces the correct result, its reasoning process is incorrect. DropLoRA and LoRA share a similar reasoning process, as do MiLoRA and PiSSA. In Table~\ref{tab:case_study1}, PiSSA’s reasoning process and outcome are both incorrect, while LoRA’s reasoning process is correct but its result is wrong. The reasoning processes and results of DropLoRA, DoRA, and MiLoRA are all correct. Note that although the reasoning processes of these three methods are correct, DropLoRA and MiLoRA’s reasoning processes explicitly highlight the keyword "least common multiple (LCM)", reflecting a more fundamental reasoning approach.

\begin{table*}[h]
\centering
\begin{tabular}{ll}
\hline
\textbf{Question} & The ratio of Mary's age to Alice's age is \(3:5\). \\&Alice is 30 years old. How many years old is Mary? \\
\hline
\textbf{DropLoRA} & If Alice is 30 years old, then the ratio of Mary's age to Alice's age is \(3:5\). \\&This means that for every 3 years old Mary is, Alice is 5 years old. \\&Since Alice is 30 years old, we can set up the equation \(\frac{3}{5} \cdot 30 = \boxed{18}\). \\&Therefore, Mary is 18 years old. The answer is: 18. \\

\hline
\textbf{DoRA} & If the ratio of Mary's age to Alice's age is \(3:5\), \\& then we can write this as \(3x:5x\), where \(x\) is a constant. \\&We are given that Alice is 30 years old, \\&so we can substitute this into the ratio to get \(3x:5x = 30\). \\&To solve for \(x\), we divide both sides of the equation by 3 and 5, \\& respectively, to get \(x = 10\).\\& Therefore, Mary's age is \(3x = 3(10) = \boxed{30}\) years old. The answer is: 30. \\
\hline
\textbf{LoRA} & If Alice is 30 years old, then the ratio of Mary's age to Alice's age is \(3:5\). \\&This means that for every 3 years old Mary is, Alice is 5 years old. \\&Since Alice is 30 years old, we can set up the equation \(\frac{3}{5} \cdot 30 = \text{Mary's age}\). \\&Simplifying, we find that Mary is \(\frac{3}{5} \cdot 30 = \boxed{18}\) years old. The answer is: 18. \\

\hline
\textbf{MiLoRA:} & If Alice is 30 years old, then the ratio of Mary's age to Alice's age is \(3:5\). \\&Let's assume Mary's age is \(x\) years old. \\&According to the ratio, we have \(\frac{x}{30} = \frac{3}{5}\). Cross-multiplying, we get \(5x = 30 \cdot 3\). \\&Simplifying, we find \(5x = 90\). Dividing both sides by 5, we get \(x = \boxed{18}\). \\&Therefore, Mary is 18 years old. The answer is: 18. \\

\hline
\textbf{PiSSA} & If the ratio of Mary's age to Alice's age is \(3:5\), \\& then we can write this as \(\frac{\text{Mary's age}}{\text{Alice's age}} = \frac{3}{5}\).\\& We are given that Alice is 30 years old, \\&so we can substitute this into the equation to get \(\frac{\text{Mary's age}}{30} = \frac{3}{5}\). \\&To solve for Mary's age, we can cross-multiply to get \(5 \cdot \text{Mary's age} = 3 \cdot 30\). \\&Simplifying, we have \(5 \cdot \text{Mary's age} = 90\). \\&Dividing both sides by 5, we find that Mary's age = \(\boxed{18}\). The answer is: 18. \\

\hline
\end{tabular}
\caption{Case Study I for Math Reasoning task on LLaMA2-7B.}
\label{tab:case_study}
\end{table*}

\begin{table*}[h]
\centering
\begin{tabular}{ll}
\hline
\textbf{Question} & What is the smallest number of marbles that could be divided up either \\& into bags of $18$ marbles or into bags of $42$ marbles,\\& with no marbles left over in each case? \\
\hline
\textbf{DropLoRA} & We can find the least common multiple (LCM) of $18$ and $42$ to\\& find the smallest number that can be  divided evenly into both $18$ and $42$.\\& Prime factorizing $18$ and $42$, we have $18=2 \times 3^2$ and $42=2 \times 3 \times 7$.\\& The LCM is then $2 \times 3^2 \times 7=126$.\\& Therefore, the smallest number of marbles that can be divided up \\& into bags of $18$ or $42$ marbles is \boxed{126}.\\& The answer is: 126. \\

\hline
\textbf{DoRA} & We want to find the smallest number that is divisible by both $18$ and $42$.\\& The prime factorization of $18$ is $2 \times 3^2$, \\& and the prime factorization of $42$ is $2 \times 3 \times 7$.\\& To find the smallest number divisible by both $18$ and $42$, \\& we take the highest power of each prime that appears in either factorization.\\& So, the smallest number divisible by both $18$ and $42$ is $2 \times 3^2 \times 7 = \boxed{126}$.\\&The answer is: 126. \\

\hline

\textbf{LoRA} & We want to find the smallest number that is divisible by both $18$ and $42$.\\& The prime factorization of $18$ is $2 \times 3^2$ \\& and the prime factorization of $42$ is $2 \times 3 \times 7$.\\& To find the smallest number divisible by both $18$ and $42$,\\& we take the highest power of each prime that appears in either factorization.\\& So, the smallest number divisible by both $18$ and $42$ is $2 \times 3^2 \times 7 = \boxed{168}$. \\&The answer is: 168. \\

\hline
\textbf{MiLoRA} & We can find the smallest number of marbles by finding \\& the least common multiple (LCM) of $18$ and $42$.\\& Prime factorizing $18$ and $42$, we have $18=2 \times 3^2$ and $42=2 \times 3 \times 7$.\\& The LCM is then $2 \times 3^2 \times 7=126$.\\& Therefore, the smallest number of marbles that could be \\& divided up into bags of $18$ marbles or $42$ marbles is $\boxed{126}$.\\& The answer is: 126. \\

\hline
\textbf{PiSSA} & The smallest number of marbles that could be divided up into bags \\& of $18$ marbles is $18$, and the smallest number of marbles that \\& could be divided up into bags of $42$ marbles is $42$.\\& Therefore, the smallest number of marbles that could be divided up\\& into both bags is $18 \times 42 = \boxed{784}$.\\& The answer is: 784. \\

\hline

\end{tabular}
\caption{Case Study II for Math Reasoning task on LLaMA2-7B.}
\label{tab:case_study1}
\end{table*}

\end{document}